\documentclass[11pt,a4paper]{article}
\PassOptionsToPackage{breaklinks}{hyperref}
\usepackage[hyperref]{emnlp2020}
\usepackage{times}
\usepackage{latexsym}
\usepackage[utf8]{inputenc}
\usepackage{amsmath}
\usepackage{amssymb}
\usepackage{booktabs}
\usepackage{multirow}
\usepackage{adjustbox}
\usepackage{enumitem}
\usepackage[colorinlistoftodos,prependcaption,textsize=tiny]{todonotes}
\usepackage{graphicx}
\usepackage{subcaption}

\usepackage{color}
\usepackage[normalem]{ulem}

\newcommand{\gm}[1]{\textcolor{red}{#1}}

\newsavebox\MBox

\usepackage{microtype}

\aclfinalcopy 
\def\aclpaperid{2406} 

\title{Message Passing for Hyper-Relational Knowledge Graphs}

\author{Mikhail Galkin\textsuperscript{1,2}, Priyansh Trivedi\textsuperscript{2}, Gaurav Maheshwari\textsuperscript{2}, Ricardo Usbeck\textsuperscript{2}, Jens Lehmann\textsuperscript{2,3} \\
  \textsuperscript{1}TU Dresden, \textsuperscript{2}Fraunhofer IAIS, \textsuperscript{3}University of Bonn \\
  \texttt{\{mikhail.galkin, priyansh.trivedi, gaurav.maheshwari,} \\
  \texttt{ricardo.usbeck, jens.lehmann\}@iais.fraunhofer.de} 
  }
  
\date{}

\begin{document}
\maketitle
\begin{abstract}
Hyper-relational knowledge graphs (KGs) (e.g., Wikidata) enable associating additional key-value pairs along with the main triple to disambiguate, or restrict the validity of a fact. 
In this work, we propose a message passing based graph encoder - \textsc{StarE} capable of modeling such hyper-relational KGs. Unlike existing approaches, \textsc{StarE} can encode an arbitrary number of additional information (\emph{qualifiers}) along with the main triple while keeping the semantic roles of qualifiers and triples intact. We also demonstrate that existing benchmarks for evaluating link prediction (LP) performance on hyper-relational KGs suffer from fundamental flaws and thus develop a new Wikidata-based dataset - WD50K. Our experiments demonstrate that \textsc{StarE} based LP model outperforms existing approaches across multiple benchmarks. We also confirm that leveraging qualifiers is vital for link prediction with gains up to 25 MRR points compared to triple-based representations.

\end{abstract}

\section{Introduction}

\begin{figure}[!ht]
    \centering
    \includegraphics[width=1.0\columnwidth]{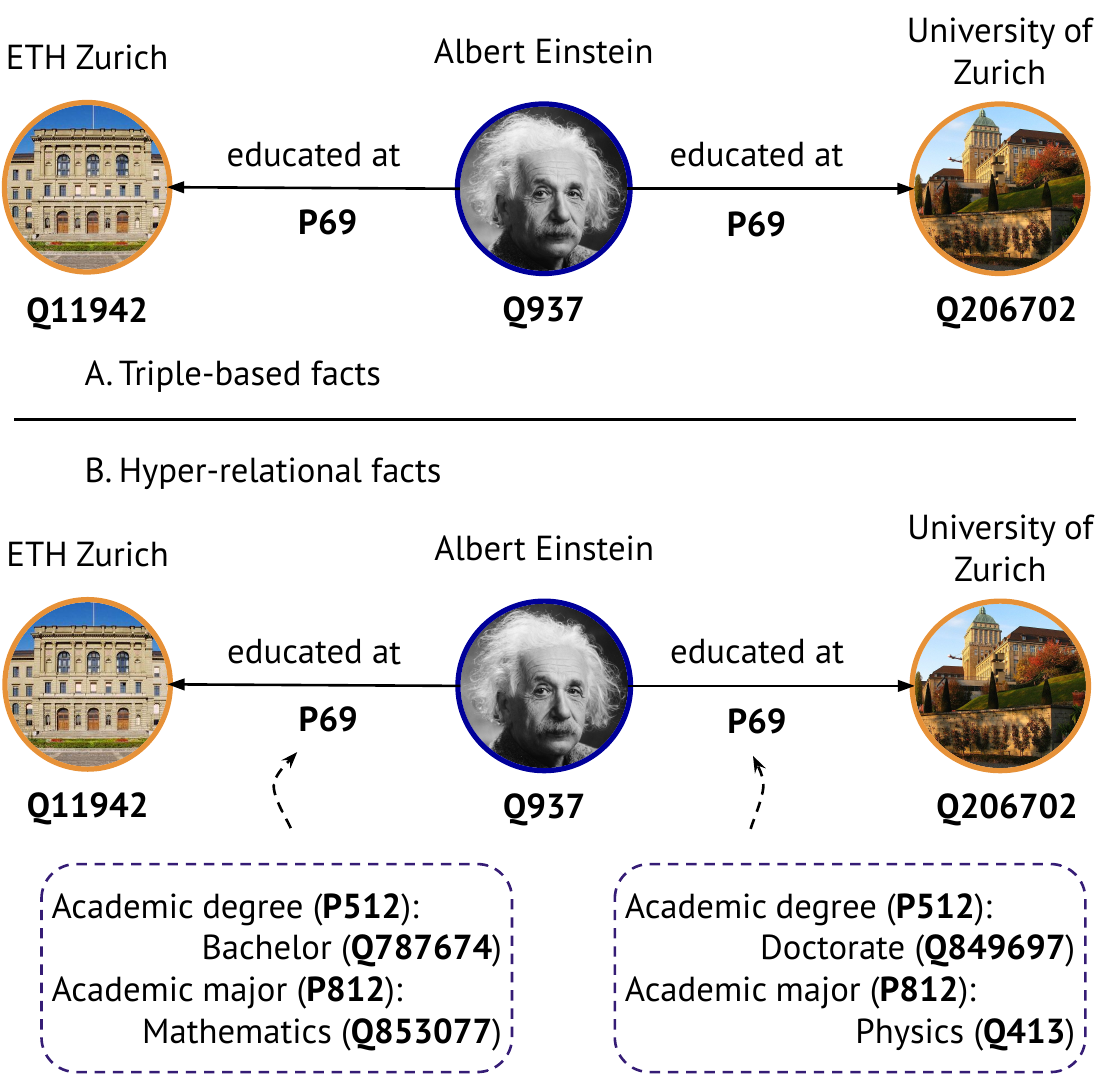}
    \caption{A comparison of triple-based and hyper-relational facts.}
    \label{fig:example}
\end{figure}

The task of link prediction over knowledge graphs (KGs) has seen a wide variety of advances over the years~\citep{DBLP:journals/corr/abs-2002-00388}. 
The objective of this task is to predict new links between entities in the graph based on the existing ones. 
A majority of these approaches are designed to work over triple-based KGs, where facts are represented as binary relations between entities.
This data model, however, doesn't allow for an intuitive representation of facts with additional information.
For instance, in Fig.~\ref{fig:example}.A, it is non-trivial to add information which can help disambiguate whether the two universities attended by \texttt{Albert Einstein} awarded him with the same degree.

This additional information can be provided in the form of key-value
restrictions over instances of binary relations between entities in recent knowledge graph models~\citep{DBLP:journals/cacm/VrandecicK14,pellissieryago,DBLP:journals/semweb/IsmayilovKALH18}. 
Such restrictions are known as \emph{qualifiers} in the Wikidata statement model~\citep{DBLP:journals/cacm/VrandecicK14} or \emph{triple metadata} in RDF*~\citep{DBLP:conf/amw/Hartig17} and RDF reification approaches~\citep{DBLP:journals/semweb/FreyMHRV19}.
These complex facts with qualifiers can be represented as \emph{hyper-relational} facts (See Sec.~\ref{sec:prelims}).
In our example (Fig.~\ref{fig:example}.B), hyper-relational facts allow to observe that \texttt{Albert Einstein} obtained different degrees at those universities.

Existing representation learning approaches for such graphs largely treat a hyper-relational fact as an n-ary (n\textgreater2) composed relation (e.g., \texttt{educatedAt\_academicDegree})~\citep{DBLP:conf/www/ZhangLMM18, 10.1145/3366423.3380188} losing entity-relation attribution; ignoring the semantic difference between a triple relation (\texttt{educatedAt}) and qualifier relation (\texttt{academicDegree})~\citep{DBLP:conf/www/GuanJWC19}, or decomposing a hyper-relational instance into multiple quintuples comprised of a triple and one qualifier key-value pair~\citep{10.1145/3366423.3380257}. 
In this work, we propose an alternate graph representation learning mechanism capable of encoding hyper-relational KGs with arbitrary number of qualifiers, while keeping the semantic roles of qualifiers and triples intact.


To accomplish this, we leverage the advances in Graph Neural Networks (GNNs), many of which are instances of the message passing~\citep{DBLP:conf/icml/GilmerSRVD17} framework, to learn latent representations of nodes and edges of a given graph. Recently, GNNs have been demonstrated~\citep{Vashishth2020Composition-based} to be capable of encoding mutli-relational (tripled based) knowledge graphs. Inspired by them, we further extend this framework to incorporate hyper-relational KGs, and propose \textsc{StarE}\footnote{The title is inspired by the RDF*~\citep{DBLP:conf/amw/Hartig17} "RDF star" proposal for standardizing hyper-relational KGs} , which to the best of our knowledge is the first GNN-based approach capable of doing so (see Sec.~\ref{sec:stare}).

Furthermore, we show that WikiPeople~\citep{DBLP:conf/www/GuanJWC19}, and JF17K~\citep{DBLP:conf/ijcai/WenLMCZ16} - two commonly used benchmarking datasets for LP over hyper-relational KGs exhibit some design flaws, which render them as ineffective benchmarks for the hyper-relational link prediction task (see Sec.~\ref{sec:WD50K}).
JF17K suffers from significant test leakage, while most of the qualifier values in WikiPeople are literals which are conventionally ignored in KG embedding approaches, rendering the dataset largely devoid of qualifiers.
Instead, we propose a new hyper-relational link prediction dataset - WD50K extracted from Wikidata~\citep{DBLP:journals/cacm/VrandecicK14} that contains statements with varying amounts of qualifiers, and use it to benchmark our approach.


Through our experiments (Sec.~\ref{sec:exp}), we find that \textsc{StarE} based model generally outperforms other approaches on the task of link prediction (LP) over hyper-relational knowledge graphs.
We provide further evidence of the fact, independent of \textsc{StarE}, that triples enriched with qualifier pairs provide additional signal beneficial for the LP task. 


\section{Related Work}



Early approaches for modelling hyper-relational graphs stem from conventional triple-based KG embedding algorithms, which often simplify complex property attributes (\emph{qualifiers}). 
For instance, m-TransH~\citep{DBLP:conf/ijcai/WenLMCZ16} requires star-to-clique conversion which results in a permanent loss of entity-relation attribution. 
Later models, e.g., RAE~\citep{DBLP:conf/www/ZhangLMM18}, HypE and HSimple introduced in~\cite{ijcai2020-303}, converted hyper-relational facts into \emph{n-ary} facts with one abstract relation 
which is supposed to loosely represent a combination of all relations of the original fact. 

Recently, GETD~\citep{10.1145/3366423.3380188} extended TuckER~\citep{DBLP:conf/emnlp/BalazevicAH19} tensor factorization approach for n-ary relational facts. 
However, the model still expects only one relation in a fact and is not able to process facts of different arity in one dataset, e.g., 3-ary and 4-ary facts have to be split and trained separately. 

NaLP~\citep{DBLP:conf/www/GuanJWC19} is a convolutional model that supports multiple entities and relations in one fact. 
However, every complex fact with $k$ qualifiers has to be broken down into $k+2$ key-value pairs with an artificial split of the main \emph{(s,p,o)} triple into $(p_s:s)$ and $(p_o:o)$ pairs. 
Consequently, all key-value pairs are treated equally thus the model does not distinguish between the main triple and relation-specific qualifiers. 

HINGE~\citep{10.1145/3366423.3380257} also adopts a convolutional framework for modeling hyper-relational facts. A main triple is iteratively convolved with every qualifier pair as a \emph{quintuple} followed by min pooling over quintuple representations. 
Although it retains the hyper-relational nature of facts, HINGE operates on a triple-quintuple level that  lacks granularity of representing a certain relation instance with its qualifiers.
Additionally, HINGE has to be trained sequentially in a \emph{curriculum learning}~\citep{10.1145/1553374.1553380} fashion requiring sorting all facts in a KG in an ascending order of the amount of qualifiers per fact which might be prohibitively expensive for large-scale graphs.

Instead, our approach directly augments a relation representation with any number of attached qualifiers properly separating auxiliary entities and relations from those in the main triple.
Additionally, we do not force any restrictions on input order of facts nor on the amount of qualifiers per fact.


Parallel to our approach are the methods that work over hypergraphs, e.g., DHNE~\citep{tu2018structural}, Hyper-SAGNN~\citep{Zhang2020Hyper-SAGNN}, and knowledge hypergraphs like HypE~\citep{ijcai2020-303}. 
We deem hyper-relational graphs and hypergraphs are conceptually different. 
As hyperedges contain multiple nodes, such hyperedges are closer to \emph{n-ary} relations \emph{$r(e_1, \dots, e_n)$} with one abstract relation. 
The attribution of entities to the main triple or qualifiers is lost, and qualifying relations are not defined. 
Combining a certain set of main and qualifying relations into one abstract $r_k()$ would lead to a combinatorial explosion of typed hyperedges since, in principle, any relation could be used in a qualifier, and there the amount of qualifiers per fact is not limited.
Therefore, modeling qualifiers in hypergraphs becomes non-trivial, and we leave such a study for future work.

\section{Preliminaries}
\label{sec:prelims}


\textbf{GNNs on Undirected Graphs:} Consider an undirected graph $\mathcal{G} = (\mathcal{V},\mathcal{E})$, where $\mathcal{V}$ represents the set of nodes  and $\mathcal{E}$ denotes the set of edges. 
Each node $v \in \mathcal{V}$ has an associated vector $\mathbf{h}_v$ and neighbourhood $\mathcal{N}(v)$. 
In the message passing framework~\citep{DBLP:conf/icml/GilmerSRVD17}, the node representations are learned iteratively via aggregating representations (messages) from their neighbors:

\begin{equation}
    \mathbf{h}_v^{k+1} = \text{UPD} \left( \mathbf{h}_v^k, \underset{u\in\mathcal{N}(v)}{\text{AGGR}} \: \phi(\mathbf{h}_v^k, \mathbf{h}_u^k, \mathbf{e}_{vu}) \right)
\end{equation}

where AGGR($\cdot$) and UPD($\cdot$) are differentiable functions for neighbourhood aggregation and node update, respectively; $\mathbf{h}_v^{(k)}$ is the representation of a node $v$ at layer $k$; $\mathbf{e}_{vu}$ is the representation of an edge between nodes $v$ and $u$. 




Different GNN architectures implement their own aggregation and update strategy. 
For example, in case of Graph Convolutional Networks (GCNs)~\citep{DBLP:conf/iclr/KipfW17} the representations of neighbours are first transformed via a weight matrix $\textbf{W}$ and then combined and passed through a non-linearity $f(\cdot)$ such as ReLU. A GCN layer $k$ can be represented as:

\begin{equation} \label{eq:GCN}
    \mathbf{h}_v^{(k)} = f \left( \sum_{u \in \mathcal{N}(v)} \mathbf{W}^{(k)} \mathbf{h}_u^{(k-1)} \right)
\end{equation}

GCN and other seminal architectures such as GAT~\citep{DBLP:conf/iclr/VelickovicCCRLB18} and GIN~\citep{DBLP:conf/iclr/XuHLJ19} do not model relation embeddings explicitly and require further modifications to support multi-relational KGs.

\textbf{GNN on Directed Multi-Relational  Graphs:} In case of a  multi-relational graph $\mathcal{G} = (\mathcal{V},\mathcal{R}, \mathcal{E})$ where $\mathcal{R}$ represents the set of relations $r$, and $\mathcal{E}$ denotes set of directed edges $(s,r,o)$ where nodes $s \in \mathcal{V}$ and $o \in \mathcal{V}$ are connected via relation $r$. The GCN formulation by~\citep{DBLP:conf/emnlp/MarcheggianiT17} assumes that the information in a directed edge flows in both directions. Thus for each edge $(s,r,o)$, an inverse edge $(o,r^{-1},s)$ is added to $\mathcal{E}$. Further, self-looping relations $(v, r^{self}, v)$, for each node $v \in \mathcal{V}$ are added to $\mathcal{E}$, enabling an update of a node state based on its previous one, and further improving normalization. 

For directed multi-relational graphs, Equation~\ref{eq:GCN} can be extended by introducing relation specific weights $\textbf{W}_r$~\citep{DBLP:conf/emnlp/MarcheggianiT17,DBLP:conf/esws/SchlichtkrullKB18}

\begin{equation}
    \mathbf{h}_v^{(k)} = f \left( \sum_{(u,r) \in \mathcal{N}(v)} \mathbf{W}_r^{(k)} \mathbf{h}_u^{(k-1)} \right)
\end{equation}

However, such networks are known to be overparameterized. 
Instead, CompGCN~\citep{Vashishth2020Composition-based} proposes to learn specific edge type vectors:

\begin{equation}
     \mathbf{h}_v^{(k)} = f \left( \sum_{(u,r) \in \mathcal{N}(v)} \mathbf{W}_{\lambda(r)}^{(k)} \phi ( \mathbf{h}_u^{(k-1)}, \mathbf{h}_r^{(k-1)} ) \right)
\label{eq:compgcn}     
\end{equation}

where $\phi(\cdot)$ is a composition function of a node $u$ with its respective relation $r$, and $\mathbf{W}_{\lambda(r)}$ is a direction-specific shared parameter for incoming, outgoing, and self-looping relations.
The composition $\phi: \mathbb{R}^d \times \mathbb{R}^d \rightarrow \mathbb{R}^d$ can be any entity-relation function akin to TransE~\citep{DBLP:conf/nips/BordesUGWY13} or DistMult~\citep{DBLP:journals/corr/YangYHGD14a}.



\textbf{Hyper-Relational Graphs:} In case of a hyper-relational graph $\mathcal{G} = (\mathcal{V}, \mathcal{R}, \mathcal{E})$, $\mathcal{E}$ is a list $(e_1, \dots, e_n)$ of edges with $e_j \in \mathcal{V} \times \mathcal{R} \times \mathcal{V} \times \mathcal{P}(\mathcal{R} \times \mathcal{V})$ for $1\le j \le n$, where $\mathcal{P}$ denotes the power set. A hyper-relational fact $e_j \in \mathcal{E}$ is usually written as a tuple $ (s,r,o,\mathcal{Q})$, where $\mathcal{Q}$ is the set of \emph{qualifier pairs} $\{(qr_{i},qv_{i}) \}$ with \emph{qualifier relations} $qr_{i} \in \mathcal{R}$ and \emph{qualifier values} $qv_{i} \in \mathcal{V}$. $(s,r,o)$ is referred to as the \emph{main triple} of the fact. We use the notation $Q_j$ to denote the qualifier pairs  
of $e_j$.
For example, under this representation scheme, one of the edges in Fig.~\ref{fig:example}.B would be (\texttt{Albert Einstein, educated at,  University of Zurich, (academic degree, Doctorate), (academic major, Physics))
}

\section{\textsc{StarE}}
\label{sec:stare}

\begin{figure}[t]
    \centering
    \includegraphics[width=\linewidth]{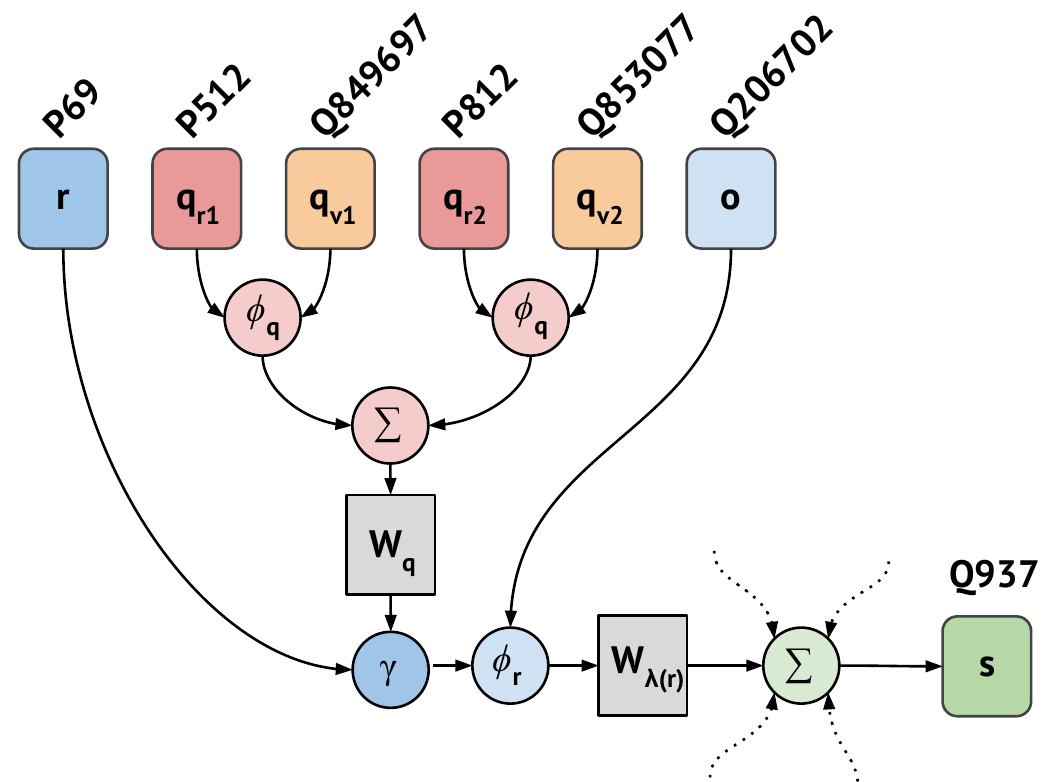}
  \caption{The mechanism in which \textsc{StarE} encodes a hyper-relational fact from Fig.~\ref{fig:example}.B. Qualifier pairs are passed through a composition function $\phi_q$, summed and transformed by $\mathbf{W}_q$. The resulting vector is then merged via $\gamma$, and $\phi_r$ with the relation and object vector, respectively. Finally, node \emph{Q937} aggregates messages from this and other hyper-relational edges.} 
  \label{fig:stare}
\end{figure}

In this section, we introduce our main contribution -- \textsc{StarE},
and show how we use it for link prediction (LP).
\textsc{StarE} (cf. Fig.~\ref{fig:stare} for the intuition) incorporates statement qualifiers $\{ (qr_{i},qv_{i}) \}$, along with the main triple $(s,r,o)$ into a message passing process.
To do this, we extend ~\autoref{eq:compgcn} by combining the edge-type embedding $\mathbf{h}_{r}$ with a fixed-length vector $\mathbf{h}_{q}$ representing qualifiers associated with a particular relation $r$ between nodes $u$ and $v$. The resultant equation is thus: 

\begin{equation}
     \mathbf{h}_v = f \left( \sum_{(u,r) \in \mathcal{N}(v)} \mathbf{W}_{\lambda(r)} \phi_r ( \mathbf{h}_u, \gamma (\mathbf{h}_r,\mathbf{h}_{q} )_{vu} ) \right)
\label{eq:stare}     
\end{equation}

where $\gamma(\cdot)$ is a function that combines the main relation representation with the representation of its qualifiers,
e.g., concatenation $[\mathbf{h}_r,\mathbf{h}_{q}]$, element-wise multiplication $\mathbf{h}_r \odot \mathbf{h}_{rq}$, or weighted sum:
\begin{equation}
     \gamma (\mathbf{h}_r,\mathbf{h}_{q} ) = \alpha \odot \mathbf{h}_r + (1-\alpha) \odot \mathbf{h}_{q}
\label{eq:stare_rel}     
\end{equation}

where $\alpha$ is a hyperparameter that controls
the flow of information from qualifier vector $\mathbf{h}_{q}$ to $\mathbf{h}_{r}$.

Finally, the qualifier vector $\mathbf{h}_{q}$ is obtained through a composition $\phi_q$ of a qualifier relation $\mathbf{h}_{qr}$ and qualifier entity $\mathbf{h}_{qv}$. 
The composition function $\phi_q$ may be any entity-relation function akin to $\phi$ (\autoref{eq:compgcn}).
The representations of different qualifier pairs are then aggregated via a position-invariant summation function and passed through a parameterized projection $\mathbf{W}_{q}$:

\begin{equation}
     \mathbf{h}_{q} = \mathbf{W}_{q} \sum_{(qr,qv) \in Q_{j_{r_{vu}}}}  \phi_q (\mathbf{h}_{qr}, \mathbf{h}_{qv})
\label{eq:stare_quals}     
\end{equation}


This formalisation allows to (i) incorporate an arbitrary number of qualifier pairs and (ii) can take into account whether entities/relations occur in the main triple or the qualifier pairs. \textsc{StarE} is the first GNN model for representation learning of hyper-relational KGs that has these characteristics.  

\begin{table*}[t]
\centering
\caption{Datasets - E in quals (R in quals) denote the amount of entities (relations) appearing only in qualifiers.}
\label{tab:datasets}
\adjustbox{max width=\linewidth}{
\begin{tabular}{lccccccccc@{}}
\toprule
\multicolumn{1}{l}{Dataset}  & Statements & w/ Quals (\%) & Entities & Relations & E in quals & R in quals & Train & Valid & Test \\ \midrule
\multicolumn{1}{l}{WD50K} & 236,507 &  32,167 (13.6\%) & 47,156 & 532 & 5460 & 45 &   166,435  & 23,913 & 46,159   \\
WD50K (33) &  102,107  &  31,866 (31.2\%) & 38,124 & 475  &  6463  &  47 & 73,406  &  10,568 &   18,133   \\
WD50K (66) &  49,167 & 31,696 (64.5\%) & 27,347 & 494  &  7167  &  53 & 35,968  &  5,154 &   8,045   \\
WD50K (100) &  31,314  &  31,314 (100\%) & 18,792 & 279  &  7862  &  75 & 22,738  &  3,279 &   5,297   \\ \midrule \midrule 
WikiPeople &  369,866  &  9,482 (2.6\%) & 34,839 & 375  & 416  &  35 & 294,439  &  37,715 &   37,712 \\
JF17K &  100,947  &  46,320 (45.9\%) & 28,645 & 322  &  3652  &  180 & 76,379  &  - &   24,568 \\ \bottomrule
\end{tabular}}
\end{table*}

\begin{figure}[t]
    \centering
    \includegraphics[width=\linewidth]{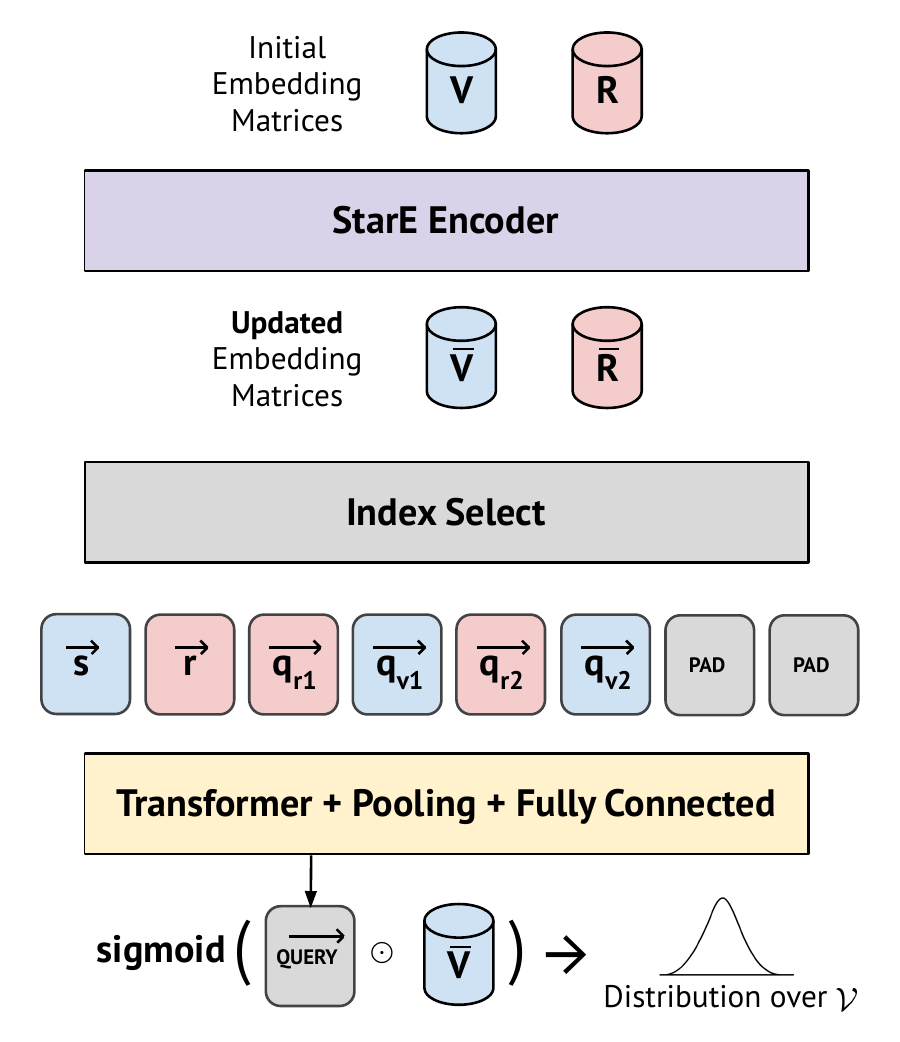}
  \caption{Architecture of a \textsc{StarE} based link prediction model. \textsc{StarE} updates the $\bar{\mathbf{V}},\bar{\mathbf{R}}$ matrices, which are then used to encode the relations in a given query before passing them through the Transformer, Pooling and fully connected layers. The fixed-dimensional output is then compared to $\bar{\mathbf{V}}$, the result of which is passed through a sigmoid function to yield a probability distribution over entities. }
  \label{fig:archi}
\end{figure}

\textbf{\textsc{StarE} for Link Prediction.}
\textsc{StarE} is a general representation learning framework for capturing the structure of hyper-relational graphs, and thus can be applied to multiple downstream tasks. In this work, we focus on LP and leave other tasks such as node classification for future work. 
In LP, given a query $(s,r,\mathcal{Q})$, the task is to predict an entity corresponding to the object position $o$.

Our link prediction model (see Fig.~\ref{fig:archi}) is composed of two parts namely (i) a \textsc{StarE} based encoder, and (b) a Transformer~\citep{DBLP:conf/nips/VaswaniSPUJGKP17} based decoder similar to CoKE~\citep{DBLP:journals/corr/abs-1911-02168}, which are jointly trained. 
We initialize two embedding matrices $\mathbf{R}, \mathbf{V}$ corresponding to relations ($\mathcal{R}$), and entities ($\mathcal{V}$) present in the dataset\footnote{As mentioned in Section~\ref{sec:prelims}, while pre-processing, we add inverse and self-loop relations to the graph. 
Note, we retain the same set of qualifiers as in the original fact while generating inverse hyper-relational facts.}. 
In every iteration, \textsc{StarE} updates the embeddings ($\bar{\mathbf{R}}, \bar{\mathbf{V}}$)  by message passing across every edge in the training set. 
In the decoding step, we first linearize the given query, and use the updated embeddings ($\bar{\mathbf{R}}, \bar{\mathbf{V}}$) to encode the entities and relations within it. 
Then, this linearized sequence is passed through the Transformer block, whose output is averaged to get a fixed-dimensional vector representation of the query.
The vector is then passed through a fully-connected layer,  multiplied with $\bar{\mathbf{V}}$ and then passed through a sigmoid, to obtain a probability distribution over all entities. 
Thereafter, it is trivial to retrieve the top $n$ candidate entities for the $o$ position in the query.

Note that we can use different decoders in this architecture. An explanation and evaluation of few decoders is provided in Appendix~\ref{app:decoders}.

\section{WD50K Dataset}
\label{sec:WD50K}
Recent approaches~\citep{DBLP:conf/www/GuanJWC19, 10.1145/3366423.3380188, 10.1145/3366423.3380257} for embedding hyper-relational KGs often use WikiPeople and JF17K as benchmarking datasets. 
We advocate that those datasets can not fully capture the task complexity.

In WikiPeople, about 13\% of statements contain at least one literal. Literals (e.g. numeric values, date-time instances or other strings, etc) in KGs are conventionally ignored~\citep{10.1145/3366423.3380257} by embedding approaches, or are incorporated through specific means~\citep{DBLP:conf/semweb/KristiadiKL0F19}. However, after removing statements with literals, less than 3\% of the remaining statements contain any qualifier pairs. Out of those, about 80\% possess only one qualifier. This fact renders WikiPeople less sensitive to hyper-relational models as performance on triple-only facts dominates the overall score.

The authors of JF17K reported\footnote{ \url{http://www.site.uottawa.ca/~yymao/JF17K/}} the dataset to contain redundant entries. In our own analysis, we detected that about 44.5\% of the test statements share the same main $(s,r,o)$ triple as the train statements.
We consider this fact as a major data leakage which allows triple-based models 
to memorize subjects and objects appearing in the test set.
 
To alleviate the above problems, we propose a new dataset, WD50K, extracted from Wikidata statements. 
The following steps are used to sample our dataset from the Wikidata RDF dump of August 2019~\footnote{\url{https://dumps.wikimedia.org/wikidatawiki/20190801/}}. 
We begin with a set of seed nodes corresponding to entities from FB15K-237 having a direct mapping in Wikidata (\texttt{P646} "Freebase ID"). 
Then, for each seed node, all statements whose \textit{main} object and \textit{qualifier} values correspond to \texttt{wikibase:Item} are extracted. 
This step results in the removal of all literals in object position. 
Similarly, all literals are filtered out from the qualifiers of the obtained statements.
To increase the connectivity in the statements graph, all the entities mentioned less than twice are dropped. 

All the statements of WD50K are randomly split into the train, test, and validation sets. 
To eliminate test set leakages we remove all statements from train and validation sets that share the same main triple \emph{(s,p,o)} with test statements. 
Finally, we remove statements from the test set that contain entities and relations not present in the train or validation sets. 
WD50K contains 236,507 statements describing 47,156 entities with 532 relations where about 14\% of statements have at least one qualifier pair.
See Table~\ref{tab:WD50K}, and Appendix~\ref{app:dataset} for further details. The dataset is publicly available\footnote{\url{https://zenodo.org/record/4036498}}.


\begin{table*}[t]
\centering
\caption{Link prediction on WikiPeople and JF17K. Results of m-TransH, RAE, NaLP-Fix and HINGE are taken from~\citep{10.1145/3366423.3380257}. Best results among hyper-relational models are in \textbf{bold}.}
\label{tab:mainexp}
\adjustbox{max width=\linewidth}{
\begin{tabular}{@{}clcccccccc@{}}
\toprule
\multirow{2}{*}{\begin{tabular}[c]{@{}c@{}}Exp\end{tabular}} & \multirow{2}{*}{Method} & \multicolumn{4}{c}{\textbf{WikiPeople}} & \multicolumn{4}{c}{\textbf{JF17K}} \\ \cmidrule(l){3-6} \cmidrule(l){7-10} 
\# &  & MRR & H@1 & H@5 & H@10 & MRR & H@1 & H@5 & H@10  \\ \midrule
1 & m-TransH & 0.063 & 0.063 & - & 0.300 & 0.206 & 0.206 & - & 0.463 \\
1 & RAE & 0.059 & 0.059 & - & 0.306 & 0.215 & 0.215 & - & 0.469 \\
1 & NaLP-Fix & 0.420 & 0.343 & - & 0.556 & 0.245 & 0.185 & - & 0.358 \\
1 & HINGE & 0.476 & \textbf{0.415} & - & 0.585 & 0.449 & 0.361 & - & 0.624 \\ \midrule \midrule
1,4 & Transformer (H) & 0.469 & 0.403 & 0.538 & 0.586 & 0.512  & 0.434 & 0.593 & 0.665 \\
1,4 & \textsc{StarE} (H) + Transformer(H) & \textbf{0.491}  & 0.398 & \textbf{0.592} & \textbf{0.648} & \textbf{0.574} & \textbf{0.496} & \textbf{0.658} & \textbf{0.725} \\ \midrule \midrule
4 &  Transformer (T) & 0.474 & 0.419 & 0.532 & 0.575 & 0.537  & 0.473 & 0.606 & 0.663 \\
4 & \textsc{StarE} (T) + Transformer (T) & 0.493 & 0.400 & 0.592 & 0.648 & 0.562 & 0.493 & 0.637 & 0.702 \\ 
\bottomrule
\end{tabular}}
\end{table*}

\section{Experiments}
\label{sec:exp}

In this section, we discuss the setup and results of multiple experiments conducted towards (i) assessing the performance of our proposed approach on the link prediction task, and (ii) analyzing the effects of including hyper-relational information during link prediction.

\subsection{Evaluating \textsc{StarE} on the LP Task}
\label{sec:exp:1}


In this experiment, we evaluate our proposed approach on the task of LP over hyper-relational graphs. 
We designed it to both compare \textsc{StarE} with the state of the art algorithms, and to better understand the contribution of the \textsc{StarE} encoder.


\textbf{Datasets:} 
We use WikiPeople\footnote{Downloaded from: \url{https://github.com/gsp2014/NaLP/tree/master/data/WikiPeople}} and JF17K\footnote{Downloaded from: \url{https://www.dropbox.com/sh/ryxohj363ujqhvq/AAAoGzAElmNnhXrWEj16UiUga?dl=0}}, despite their design flaws (see Sec.~\ref{sec:WD50K}) to illustrate the performance differences with existing approaches.
We also provide a benchmark of our approach on the WD50K dataset introduced in this article.
Note that as described by~\cite{10.1145/3366423.3380257}, we drop all statements containing literals in WikiPeople. 
Further datasets statistics are presented in Table~\ref{tab:datasets}.

\textbf{Baselines:}
In this experiment, we compare against previous hyper-relational approaches namely: (i) m-TransH~\citep{DBLP:conf/ijcai/WenLMCZ16}, ii) RAE~\citep{DBLP:conf/www/ZhangLMM18}, (iii) NaLP-Fix (an improved version of NaLP~\citep{DBLP:conf/www/GuanJWC19} as proposed in~\citep{10.1145/3366423.3380257}), and (iv) HINGE~\citep{10.1145/3366423.3380257}.

To assess the significance of the \textsc{StarE} encoder, we also train a simpler model where the Transformer based decoder directly uses the randomly initialized embedding matrices without the \textsc{StarE} encoder.
We call this model \textbf{Transformer (H)}, and the one with the \textsc{StarE} encoder \textbf{\textsc{StarE} (H) + Transformer (H)}. Here (H) represents that the input to the model is a hyper-relational fact. Later, we also experiment with triples as input and represent them with (T) (see Sec.~\ref{exp:triples}).

\textbf{Evaluation:} For all the systems discussed above, we report various performance metrics when predicting the subject and object of hyper-relational facts.
We adopt the \emph{filtered} setting introduced in~\citep{DBLP:conf/nips/BordesUGWY13} for computing mean reciprocal rank (MRR) and hits at 1, 5, and 10 (H@1, H@5, H@10). 
The metrics are computed for subject and object prediction separately and are then averaged. 




\textbf{Training:}
We train the model in 1-N setting using binary cross entropy loss with label smoothing as in~\citep{DBLP:conf/aaai/DettmersMS018,Vashishth2020Composition-based} with Adam~\citep{DBLP:journals/corr/KingmaB14} optimizer for 500 epochs on WikiPeople and for 400 epochs on JF17K and WD50K datasets. 
Hyperparameters were selected by manual fine tuning with further details in Appendix~\ref{app:hyperparams}.
\textsc{StarE} is implementated with PyTorch Geometric~\citep{DBLP:journals/corr/abs-1903-02428} and is publicly available here\footnote{\url{https://github.com/migalkin/StarE}}.


\begin{table*}[t]
\centering
\caption{Link prediction on WD50K graphs with different ratio of qualifiers. Best results are in \textbf{bold}.}
\label{tab:WD50K}
\adjustbox{max width=\linewidth}{
\begin{tabular}{@{}clcccccccccccc@{}}
\toprule
\multirow{2}{*}{\begin{tabular}[c]{@{}c@{}}Exp\\ \#\end{tabular}} & Dataset $\rightarrow$ & \multicolumn{3}{c}{\textbf{WD50K}} & \multicolumn{3}{c}{\textbf{WD50K (33)}} & \multicolumn{3}{c}{\textbf{WD50K (66)}} & \multicolumn{3}{c}{\textbf{WD50K (100)}} \\ \cmidrule(l){3-5} \cmidrule(l){6-8} \cmidrule(l){9-11} \cmidrule(l){12-14}
 & Method $\downarrow$ & MRR & H@1 & \multicolumn{1}{c}{H@10} & MRR & H@1 & \multicolumn{1}{c}{H@10} & MRR & H@1 & \multicolumn{1}{c}{H@10} & MRR & H@1 & H@10 \\ \midrule
4 & Baseline (Transformer (T)) & 0.275 & 0.207 & 0.404 & 0.218 & 0.158 & 0.334 & 0.270 & 0.197 & 0.417 & 0.351 & 0.261 & 0.530 \\ 
4 & \textsc{StarE} (T) + Transformer(T) & 0.308 & 0.228 & 0.465 & 0.246 & 0.173 & 0.388 & 0.297 & 0.212 & 0.470 & 0.380 & 0.276 & 0.584 \\
\midrule \midrule
4 & NaLP-Fix & 0.177 & 0.131 & 0.264 & 0.204 & 0.164 & 0.277 & 0.334 & 0.284 & 0.423 & 0.458 & 0.398 & 0.563 \\
4 & HINGE & 0.243 & 0.176 & 0.377 & 0.253 & 0.190 & 0.372 & 0.378 & 0.307 & 0.512 & 0.492 & 0.417 & 0.636 \\
1,2,4 & Baseline (Transformer (H)) & 0.286 & 0.222 & 0.406 & 0.276 & 0.227 & 0.371 & 0.404 & 0.352 & 0.502 & 0.562 & 0.499 & 0.677 \\

1,2,4 & \textsc{StarE} (H) + Transformer(H)  & 
\textbf{0.349} & \textbf{0.271} & \textbf{0.496} & \textbf{0.331} & \textbf{0.268} & \textbf{0.451} & \textbf{0.481} & \textbf{0.420} & \textbf{0.594} & \textbf{0.654} & \textbf{0.588} & \textbf{0.777} \\
\bottomrule
\end{tabular}}
\end{table*}

\textbf{Results and Discussion:} 
The results of this experiment can be found in Table~\ref{tab:mainexp}.
We observe that the \textsc{StarE}  encoder based model outperforms the other hyper-relational models across  WikiPeople and JF17K.
On JF17K, \textsc{StarE} (H) + Transformer (H) reports a gain of 11.3 (25\%) MRR points, 13 (33\%) H@1, and 7.8 (12\%) H@10 points when compared to the next-best approach.
Recall that JF17K suffers from a major test set leakage (Sec.~\ref{sec:WD50K}), which we investigate in greater detail in Exp. 4 (Sec.~\ref{exp:triples}) below. On WikiPeople, HINGE has a higher H@1 score than \textsc{StarE} (H) + Transformer (H). However, its H@10 is lower than H@5 of our approach, i.e., top five predictions of the \textsc{StarE} model are more likely to contain a correct answer than top 10 predictions of HINGE. We can thus claim our \textsc{StarE} based model to be competitive with, if not outperforming  the state of the art on the task of link prediction over hyper-relational KGs, albeit on less-than-ideal baselines.

We further present the performance of our approach as a  baseline on the WD50K dataset in Table~\ref{tab:WD50K}. With an MRR score of 0.349, H@1 of 0.271, and H@10 of 0.496, we find that the task is far from solved, 
however, the \textsc{StarE}-based approaches provide effective, non-trivial baselines. 

Note that Transformer (H) (without \textsc{StarE}) also performs competitively to HINGE. This suggests that the aforementioned gains in metrics of our approach 
cannot all be attributed to \textsc{StarE}'s innate ability to effectively encode the hyper-relational information. That said, upon comparing the performance of \textsc{StarE} (H) + Transformer (H) and Transformer (H), we find that using \textsc{StarE} is consistently advantageous across all the datasets. 

\subsection{Impact of Ratio of Statements with and Without Qualifier Pairs}
\label{exp:quals}

Based on the relatively high performance of Transformer (H) (without the encoder) in the previous experiment,  we study the relationship between the amount of hyper-relational information (qualifiers), and the ability of \textsc{StarE} to incorporate it for the LP task. 
Here, we sample datasets from WD50K, with varying ratio of facts with qualifier pairs to the total number of facts in the KG.
Specifically, we sample three datasets namely, \emph{WD50K (33)}, \emph{WD50K (66)}, and \emph{WD50k (100)} containing approximately 33\%, 66\%, and 100\% of such hyper-relational facts, respectively. We use the same experimental setup as the one discussed in the previous section. Table~\ref{tab:WD50K} presents the result of this experiment.

We observe that across all metrics, \textsc{StarE} (H) + Transformer (H) performs increasingly better than Transformer (H), as the ratio of qualifier pairs increases in the dataset.
Concretely, the difference in their H@1 scores is 4.1, 6.8, and 8.9 points on \emph{WD50K (33)}, \emph{WD50K (66)}, and \emph{WD50K (100)} respectively.
These and the Sec.~\ref{sec:exp:1} results confirm that 
\textsc{StarE} is better suited to utilize the qualifier information available in the KG, (ii) which when leveraged by a transformer decoder, outperforms other hyper-relational LP approaches, and (iii) that \textsc{StarE}'s positive effects increases as the amount of qualifiers in the task increases.



\begin{figure*}[t]
    \centering
    \includegraphics[width=\linewidth]{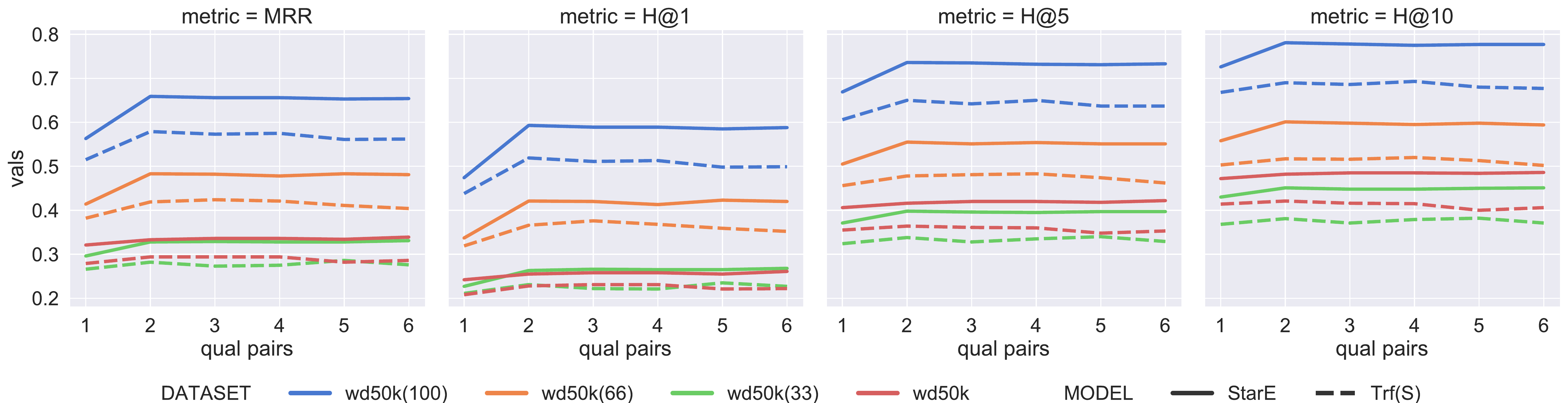}
    \caption{Statement length experiment. \textsc{StarE} (H) + Transformer (H) saturates after two qualifiers with slightly increase, whereas Transformer (H) is unstable in handling qualifiers.}
    \label{fig:WD50K_slen}
\end{figure*}

\subsection{Impact of Number of Qualifiers per Statement}
\label{exp:statement_len}

In WD50K, as in Wikidata, the number of qualifiers corresponding to a statement varies significantly (see Appendix~\ref{app:dataset}).
In this experiment, we intend to quantify its effect on the model performance.

To do so, we create multiple variants of WD50K, each containing statements with up to $n$ qualifiers($n \in [1, 6]$).
In other words, for a given number $n$, we collect all the statements which have less than $n$ qualifiers. 
If a statement contains more than $n$ qualifiers, we arbitrarily choose $n$ qualifiers amongst them. 
Thus, the total number of facts remains the same across these variants.
Figure~\ref{fig:WD50K_slen} presents the result of this experiment.

For all the datasets, we find that two qualifier pairs are enough for our model performance to saturate. 
This might be an attribute of the underlying characteristic of the dataset or the model's inability to aggregate information from longer statements. 
We leave the further analysis of this for the future work.
However, we observe that in case of WD50K and other datasets, \textsc{StarE (H)} + Transformer (H) slightly improves or remains stable with increase of statement length, while Transformer (H) shows degradation in performance. 


\subsection{Comparison to Triple Baselines}
\label{exp:triples}

To further understand the role of qualifier information in the LP task, we design an experiment to gauge the performance difference between models on hyper-relational KG and triple-based KG. Concretely, we create a new triple-only dataset by pruning all qualifier information from the statements in WikiPeople, JF17K, and WD50K.
That is, two statements that describe the same main fact $(s, r, o, \{(qr_1,  qv_1), (qr_2, qv_2)\}$ and $(s, r, o, \{(qr_3, qv_3), (qr_4, qv_4)\})$ are reduced to one triple $(s, r, o)$.
Thus, the overall amount of distinct entities and relations is reduced\gm{,} but the amount of subjects and objects in \emph{main triples} for the LP task is the same.

We introduce \textbf{\textsc{StarE} (T) + Transformer (T)}, a model for this experiment.
\textsc{StarE} (T) is similar to CompGCN~\citep{Vashishth2020Composition-based}, and can only model triple-based $(s, r, o)$ facts.
Since inputs to the Transformer decoder are linearized queries, we can trivially implement Transformer (T) by ignoring qualifier pairs during this linearization. 
The results are available in Table~\ref{tab:mainexp}, and Table~\ref{tab:WD50K}.

We observe that triple-only baselines yield competitive results on JF17K and WikiPeople compared to hyper-relational models (See Table~\ref{tab:mainexp}). 
As WikiPeople contains less than 3\% of hyper-relational facts, the biggest contribution to the overall performance is dominated by the triple-only performance.
We attribute the strong performance of the triple-only baseline on JF17K to the identified data leakage pertaining to this dataset. 
In other words, JF17K in its hyper-relational form exhibits similar issues identified by~\citep{DBLP:journals/corr/abs-2003-08001} as in FB15k and WN18 datasets proposed in~\citep{DBLP:conf/nips/BordesUGWY13} for triple-based LP task.
We thus perform another experiment after cleaning JF17K from the assumed data leakage and report the results in Table~\ref{tab:jf17k_clean} below.

\begin{table}[!h]
\centering
\caption{StarE (H) + Transformer (H) denoted as (H) and Transformer (T) as (T) on the original JF17K and cleaned JF17K}
\label{tab:jf17k_clean}
\begin{tabular}{@{}lcccc@{}}
\toprule
\multirow{2}{*}{} & \multicolumn{2}{c}{\textbf{JF17K (original)}} & \multicolumn{2}{c}{\textbf{JF17K (cleaned)}} \\ \cmidrule(l){2-3} \cmidrule(l){4-5}
 & H & T & H & T \\ \midrule
MRR & 0.574 & 0.534 & 0.376 & 0.334 \\
H@1 & 0.496 & 0.471 & 0.278 & 0.242 \\
H@5 & 0.658 & 0.602 &  0.485 & 0.428 \\
H@10 & 0.725 & 0.661 &  0.582  &  0.514 \\ \bottomrule
\end{tabular}
\end{table}

We observe a drastic performance drop of about 20 MRR points in both models which provide experimental evidence of the flaws discussed in Sec.~\ref{sec:WD50K}. 
We encourage future works in this domain to refrain from using these datasets in experiments.

In the case of WD50K (where about 13\% of facts have qualifiers) the \textsc{StarE} (H) + Transformer (H) yields about 16\%, 23\%, and 11\% of relative improvement over the best performing triple-only baseline across MRR, H@1 and H@10, respectively (see Table~\ref{tab:WD50K}). Akin to the previous experiment, we observe that increasing the ratio of hyper-relational facts in the dataset leads to even higher performance boosts. 
In particular, on \emph{WD50K (100)}, the H@1 of our hyper-relational model is higher than the H@10 of the triple baseline.
This difference corresponds to 30 MRR and 32 H@1 points which is about 85\% and 123\% relative improvement, respectively.

Based on the above observations we therefore conclude, that information in hyper-relational facts indeed helps to better predict subjects and objects in the main triples of those facts.

\section{Conclusion}

We presented \textsc{StarE}, an instance of the message passing framework for representation learning over hyper-relational KGs. 
Experimental results suggest that \textsc{StarE} performs competitively on link prediction tasks over existing hyper-relational approaches and greatly outperforms triple-only baselines. 
In the future, we aim at applying \textsc{StarE} for node and graph classification tasks as well as extend our approach to large-scale KGs.

We also identified significant flaws in existing link prediction datasets and proposed WD50K, a novel, Wikidata-based hyper-relational dataset that is closer to real-world graphs and better captures the complexity of the link prediction task.  
In the future, we plan to enrich WD50K entities with class labels and probe it against node classification tasks.



\section*{Acknowledgments}
We thank the Center for Information Services and High Performance Computing (ZIH) at TU Dresden for generous allocations of computer time. We acknowledge the support of the following projects: SPEAKER (FKZ 01MK20011A), JOSEPH (Fraunhofer Zukunftsstiftung), H2020 Cleopatra (GA 812997), ML2R (FKZ 01 15 18038 A/B/C), MLwin (01IS18050 D/F), ScaDS (01IS18026A).

\bibliographystyle{acl_natbib}
\bibliography{emnlp2020}

\begin{thebibliography}{43}
\expandafter\ifx\csname natexlab\endcsname\relax\def\natexlab#1{#1}\fi

\bibitem[{Akrami et~al.(2020)Akrami, Saeef, Zhang, Hu, and
  Li}]{DBLP:journals/corr/abs-2003-08001}
Farahnaz Akrami, Mohammed~Samiul Saeef, Qingheng Zhang, Wei Hu, and Chengkai
  Li. 2020.
\newblock Realistic re-evaluation of knowledge graph completion methods: An
  experimental study.
\newblock \emph{CoRR}, abs/2003.08001.

\bibitem[{Balazevic et~al.(2019)Balazevic, Allen, and
  Hospedales}]{DBLP:conf/emnlp/BalazevicAH19}
Ivana Balazevic, Carl Allen, and Timothy~M. Hospedales. 2019.
\newblock Tucker: Tensor factorization for knowledge graph completion.
\newblock In \emph{{EMNLP-IJCNLP} 2019}, pages 5184--5193.

\bibitem[{Bengio et~al.(2009)Bengio, Louradour, Collobert, and
  Weston}]{10.1145/1553374.1553380}
Yoshua Bengio, J\'{e}r\^{o}me Louradour, Ronan Collobert, and Jason Weston.
  2009.
\newblock Curriculum learning.
\newblock In \emph{Proceedings of the 26th Annual International Conference on
  Machine Learning}, page 41–48.

\bibitem[{Bordes et~al.(2013)Bordes, Usunier, Garc{\'{\i}}a{-}Dur{\'{a}}n,
  Weston, and Yakhnenko}]{DBLP:conf/nips/BordesUGWY13}
Antoine Bordes, Nicolas Usunier, Alberto Garc{\'{\i}}a{-}Dur{\'{a}}n, Jason
  Weston, and Oksana Yakhnenko. 2013.
\newblock Translating embeddings for modeling multi-relational data.
\newblock In \emph{Advances in Neural Information Processing Systems}, pages
  2787--2795.

\bibitem[{Chakraborty et~al.(2019)Chakraborty, Lukovnikov, Maheshwari, Trivedi,
  Lehmann, and Fischer}]{DBLP:journals/corr/abs-1907-09361}
Nilesh Chakraborty, Denis Lukovnikov, Gaurav Maheshwari, Priyansh Trivedi, Jens
  Lehmann, and Asja Fischer. 2019.
\newblock Introduction to neural network based approaches for question
  answering over knowledge graphs.
\newblock \emph{CoRR}, abs/1907.09361.

\bibitem[{Cohen et~al.(2020)Cohen, Sun, Hofer, and Siegler}]{Cohen2020Scalable}
William~W. Cohen, Haitian Sun, R.~Alex Hofer, and Matthew Siegler. 2020.
\newblock Scalable neural methods for reasoning with a symbolic knowledge base.
\newblock In \emph{International Conference on Learning Representations}.

\bibitem[{Dettmers et~al.(2018)Dettmers, Minervini, Stenetorp, and
  Riedel}]{DBLP:conf/aaai/DettmersMS018}
Tim Dettmers, Pasquale Minervini, Pontus Stenetorp, and Sebastian Riedel. 2018.
\newblock Convolutional 2d knowledge graph embeddings.
\newblock In \emph{Proceedings of the Thirty-Second {AAAI} Conference on
  Artificial Intelligence}, pages 1811--1818.

\bibitem[{Dubey et~al.(2019)Dubey, Banerjee, Abdelkawi, and
  Lehmann}]{dubey2019lc}
Mohnish Dubey, Debayan Banerjee, Abdelrahman Abdelkawi, and Jens Lehmann. 2019.
\newblock Lc-quad 2.0: a large dataset for complex question answering over
  wikidata and dbpedia.
\newblock In \emph{International Semantic Web Conference}.

\bibitem[{Fatemi et~al.(2020)Fatemi, Taslakian, Vazquez, and
  Poole}]{ijcai2020-303}
Bahare Fatemi, Perouz Taslakian, David Vazquez, and David Poole. 2020.
\newblock Knowledge hypergraphs: Prediction beyond binary relations.
\newblock In \emph{Proceedings of the Twenty-Ninth International Joint
  Conference on Artificial Intelligence, {IJCAI-20}}, pages 2191--2197.

\bibitem[{Fey and Lenssen(2019)}]{DBLP:journals/corr/abs-1903-02428}
Matthias Fey and Jan~Eric Lenssen. 2019.
\newblock Fast graph representation learning with pytorch geometric.
\newblock \emph{CoRR}, abs/1903.02428.

\bibitem[{Frey et~al.(2019)Frey, M{\"{u}}ller, Hellmann, Rahm, and
  Vidal}]{DBLP:journals/semweb/FreyMHRV19}
Johannes Frey, Kay M{\"{u}}ller, Sebastian Hellmann, Erhard Rahm, and
  Maria{-}Esther Vidal. 2019.
\newblock Evaluation of metadata representations in {RDF} stores.
\newblock \emph{Semantic Web}, 10(2):205--229.

\bibitem[{Gilmer et~al.(2017)Gilmer, Schoenholz, Riley, Vinyals, and
  Dahl}]{DBLP:conf/icml/GilmerSRVD17}
Justin Gilmer, Samuel~S. Schoenholz, Patrick~F. Riley, Oriol Vinyals, and
  George~E. Dahl. 2017.
\newblock Neural message passing for quantum chemistry.
\newblock In \emph{Proceedings of the 34th International Conference on Machine
  Learning, {ICML} 2017, Sydney, NSW, Australia, 6-11 August 2017}, volume~70
  of \emph{Proceedings of Machine Learning Research}, pages 1263--1272. {PMLR}.

\bibitem[{Guan et~al.(2019)Guan, Jin, Wang, and
  Cheng}]{DBLP:conf/www/GuanJWC19}
Saiping Guan, Xiaolong Jin, Yuanzhuo Wang, and Xueqi Cheng. 2019.
\newblock Link prediction on n-ary relational data.
\newblock In \emph{The World Wide Web Conference, {WWW} 2019}, pages 583--593.

\bibitem[{Hartig(2017)}]{DBLP:conf/amw/Hartig17}
Olaf Hartig. 2017.
\newblock Foundations of rdf{\(\star\)} and sparql{\(\star\)} (an alternative
  approach to statement-level metadata in {RDF)}.
\newblock In \emph{Proceedings of the 11th Alberto Mendelzon International
  Workshop on Foundations of Data Management and the Web}.

\bibitem[{Hayashi et~al.(2019)Hayashi, Hu, Xiong, and
  Neubig}]{DBLP:journals/corr/abs-1908-07690}
Hiroaki Hayashi, Zecong Hu, Chenyan Xiong, and Graham Neubig. 2019.
\newblock \href {http://arxiv.org/abs/1908.07690} {Latent relation language
  models}.
\newblock \emph{CoRR}, abs/1908.07690.

\bibitem[{Ismayilov et~al.(2018)Ismayilov, Kontokostas, Auer, Lehmann, and
  Hellmann}]{DBLP:journals/semweb/IsmayilovKALH18}
Ali Ismayilov, Dimitris Kontokostas, S{\"{o}}ren Auer, Jens Lehmann, and
  Sebastian Hellmann. 2018.
\newblock Wikidata through the eyes of dbpedia.
\newblock \emph{Semantic Web}, 9(4):493--503.

\bibitem[{Ji et~al.(2020)Ji, Pan, Cambria, Marttinen, and
  Yu}]{DBLP:journals/corr/abs-2002-00388}
Shaoxiong Ji, Shirui Pan, Erik Cambria, Pekka Marttinen, and Philip~S. Yu.
  2020.
\newblock A survey on knowledge graphs: Representation, acquisition and
  applications.
\newblock \emph{CoRR}, abs/2002.00388.

\bibitem[{Kingma and Ba(2015)}]{DBLP:journals/corr/KingmaB14}
Diederik~P. Kingma and Jimmy Ba. 2015.
\newblock Adam: {A} method for stochastic optimization.
\newblock In \emph{3rd International Conference on Learning Representations,
  {ICLR} 2015}.

\bibitem[{Kipf and Welling(2017)}]{DBLP:conf/iclr/KipfW17}
Thomas~N. Kipf and Max Welling. 2017.
\newblock Semi-supervised classification with graph convolutional networks.
\newblock In \emph{5th International Conference on Learning Representations,
  {ICLR} 2017, Toulon, France, April 24-26, 2017, Conference Track
  Proceedings}.

\bibitem[{Kristiadi et~al.(2019)Kristiadi, Khan, Lukovnikov, Lehmann, and
  Fischer}]{DBLP:conf/semweb/KristiadiKL0F19}
Agustinus Kristiadi, Mohammad~Asif Khan, Denis Lukovnikov, Jens Lehmann, and
  Asja Fischer. 2019.
\newblock Incorporating literals into knowledge graph embeddings.
\newblock In \emph{The Semantic Web - {ISWC} 2019}, volume 11778 of
  \emph{Lecture Notes in Computer Science}, pages 347--363. Springer.

\bibitem[{Liu et~al.(2020)Liu, Yao, and Li}]{10.1145/3366423.3380188}
Yu~Liu, Quanming Yao, and Yong Li. 2020.
\newblock Generalizing tensor decomposition for n-ary relational knowledge
  bases.
\newblock In \emph{Proceedings of The Web Conference 2020}, page 1104–1114.

\bibitem[{Marcheggiani and Titov(2017)}]{DBLP:conf/emnlp/MarcheggianiT17}
Diego Marcheggiani and Ivan Titov. 2017.
\newblock Encoding sentences with graph convolutional networks for semantic
  role labeling.
\newblock In \emph{Proceedings of the 2017 Conference on Empirical Methods in
  Natural Language Processing, {EMNLP} 2017, Copenhagen, Denmark, September
  9-11, 2017}, pages 1506--1515.

\bibitem[{Mesquita et~al.(2019)Mesquita, Cannaviccio, Schmidek, Mirza, and
  Barbosa}]{DBLP:conf/emnlp/MesquitaCSMB19}
Filipe Mesquita, Matteo Cannaviccio, Jordan Schmidek, Paramita Mirza, and
  Denilson Barbosa. 2019.
\newblock Knowledgenet: {A} benchmark dataset for knowledge base population.
\newblock In \emph{{EMNLP-IJCNLP} 2019}, pages 749--758. Association for
  Computational Linguistics.

\bibitem[{Nguyen et~al.(2018)Nguyen, Nguyen, Nguyen, and
  Phung}]{DBLP:conf/naacl/NguyenNNP18}
Dai~Quoc Nguyen, Tu~Dinh Nguyen, Dat~Quoc Nguyen, and Dinh~Q. Phung. 2018.
\newblock A novel embedding model for knowledge base completion based on
  convolutional neural network.
\newblock In \emph{Proceedings of the 2018 Conference of the North American
  Chapter of the Association for Computational Linguistics: Human Language
  Technologies, NAACL-HLT}, pages 327--333.

\bibitem[{Nickel et~al.(2016)Nickel, Rosasco, and
  Poggio}]{DBLP:conf/aaai/NickelRP16}
Maximilian Nickel, Lorenzo Rosasco, and Tomaso~A. Poggio. 2016.
\newblock Holographic embeddings of knowledge graphs.
\newblock In \emph{Proceedings of the Thirtieth {AAAI} Conference on Artificial
  Intelligence}. {AAAI} Press.

\bibitem[{Pellissier-Tanon et~al.(2020)Pellissier-Tanon, Weikum, and
  Suchanek}]{pellissieryago}
Thomas Pellissier-Tanon, Gerhard Weikum, and Fabian Suchanek. 2020.
\newblock Yago 4: A reason-able knowledge base.
\newblock In \emph{Extended Semantic Web Conference, {ESWC} 2020}.

\bibitem[{Rosso et~al.(2020)Rosso, Yang, and
  Cudr\'{e}-Mauroux}]{10.1145/3366423.3380257}
Paolo Rosso, Dingqi Yang, and Philippe Cudr\'{e}-Mauroux. 2020.
\newblock Beyond triplets: Hyper-relational knowledge graph embedding for link
  prediction.
\newblock In \emph{Proceedings of The Web Conference 2020}, page 1885–1896.

\bibitem[{Schlichtkrull et~al.(2018)Schlichtkrull, Kipf, Bloem, van~den Berg,
  Titov, and Welling}]{DBLP:conf/esws/SchlichtkrullKB18}
Michael~Sejr Schlichtkrull, Thomas~N. Kipf, Peter Bloem, Rianne van~den Berg,
  Ivan Titov, and Max Welling. 2018.
\newblock Modeling relational data with graph convolutional networks.
\newblock In \emph{The Semantic Web - 15th International Conference, {ESWC}
  2018, Heraklion, Crete, Greece, June 3-7, 2018, Proceedings}, pages 593--607.

\bibitem[{Sun et~al.(2019)Sun, Deng, Nie, and Tang}]{DBLP:conf/iclr/SunDNT19}
Zhiqing Sun, Zhi{-}Hong Deng, Jian{-}Yun Nie, and Jian Tang. 2019.
\newblock Rotate: Knowledge graph embedding by relational rotation in complex
  space.
\newblock In \emph{7th International Conference on Learning Representations,
  {ICLR} 2019}.

\bibitem[{Toutanova and Chen(2015)}]{toutanova2015observed}
Kristina Toutanova and Danqi Chen. 2015.
\newblock Observed versus latent features for knowledge base and text
  inference.
\newblock In \emph{Proceedings of the 3rd Workshop on Continuous Vector Space
  Models and their Compositionality}, pages 57--66.

\bibitem[{Tu et~al.(2018)Tu, Cui, Wang, Wang, and Zhu}]{tu2018structural}
Ke~Tu, Peng Cui, Xiao Wang, Fei Wang, and Wenwu Zhu. 2018.
\newblock Structural deep embedding for hyper-networks.
\newblock In \emph{Proceedings of the 23rd AAAI Conference on Artificial
  Intelligence}.

\bibitem[{Vashishth et~al.(2020)Vashishth, Sanyal, Nitin, and
  Talukdar}]{Vashishth2020Composition-based}
Shikhar Vashishth, Soumya Sanyal, Vikram Nitin, and Partha Talukdar. 2020.
\newblock Composition-based multi-relational graph convolutional networks.
\newblock In \emph{International Conference on Learning Representations}.

\bibitem[{Vaswani et~al.(2017)Vaswani, Shazeer, Parmar, Uszkoreit, Jones,
  Gomez, Kaiser, and Polosukhin}]{DBLP:conf/nips/VaswaniSPUJGKP17}
Ashish Vaswani, Noam Shazeer, Niki Parmar, Jakob Uszkoreit, Llion Jones,
  Aidan~N. Gomez, Lukasz Kaiser, and Illia Polosukhin. 2017.
\newblock Attention is all you need.
\newblock In \emph{Advances in Neural Information Processing Systems {NIPS}
  2017}, pages 5998--6008.

\bibitem[{Velickovic et~al.(2018)Velickovic, Cucurull, Casanova, Romero,
  Li{\`{o}}, and Bengio}]{DBLP:conf/iclr/VelickovicCCRLB18}
Petar Velickovic, Guillem Cucurull, Arantxa Casanova, Adriana Romero, Pietro
  Li{\`{o}}, and Yoshua Bengio. 2018.
\newblock Graph attention networks.
\newblock In \emph{6th International Conference on Learning Representations,
  {ICLR}}.

\bibitem[{Vrandecic and Kr{\"{o}}tzsch(2014)}]{DBLP:journals/cacm/VrandecicK14}
Denny Vrandecic and Markus Kr{\"{o}}tzsch. 2014.
\newblock Wikidata: a free collaborative knowledgebase.
\newblock \emph{Commun. {ACM}}, 57(10):78--85.

\bibitem[{Wang et~al.(2019{\natexlab{a}})Wang, Huang, Wang, Dai, Jiang, Liu,
  Lyu, Zhu, and Wu}]{DBLP:journals/corr/abs-1911-02168}
Quan Wang, Pingping Huang, Haifeng Wang, Songtai Dai, Wenbin Jiang, Jing Liu,
  Yajuan Lyu, Yong Zhu, and Hua Wu. 2019{\natexlab{a}}.
\newblock Coke: Contextualized knowledge graph embedding.
\newblock \emph{CoRR}, abs/1911.02168.

\bibitem[{Wang et~al.(2019{\natexlab{b}})Wang, Gao, Zhu, Liu, Li, and
  Tang}]{DBLP:journals/corr/abs-1911-06136}
Xiaozhi Wang, Tianyu Gao, Zhaocheng Zhu, Zhiyuan Liu, Juanzi Li, and Jian Tang.
  2019{\natexlab{b}}.
\newblock {KEPLER:} {A} unified model for knowledge embedding and pre-trained
  language representation.
\newblock \emph{CoRR}, abs/1911.06136.

\bibitem[{Wen et~al.(2016)Wen, Li, Mao, Chen, and
  Zhang}]{DBLP:conf/ijcai/WenLMCZ16}
Jianfeng Wen, Jianxin Li, Yongyi Mao, Shini Chen, and Richong Zhang. 2016.
\newblock On the representation and embedding of knowledge bases beyond binary
  relations.
\newblock In \emph{Proceedings of the Twenty-Fifth International Joint
  Conference on Artificial Intelligence, {IJCAI} 2016}, pages 1300--1307.

\bibitem[{Xiong et~al.(2020)Xiong, Du, Wang, and
  Stoyanov}]{Xiong2020Pretrained}
Wenhan Xiong, Jingfei Du, William~Yang Wang, and Veselin Stoyanov. 2020.
\newblock Pretrained encyclopedia: Weakly supervised knowledge-pretrained
  language model.
\newblock In \emph{International Conference on Learning Representations}.

\bibitem[{Xu et~al.(2019)Xu, Hu, Leskovec, and
  Jegelka}]{DBLP:conf/iclr/XuHLJ19}
Keyulu Xu, Weihua Hu, Jure Leskovec, and Stefanie Jegelka. 2019.
\newblock How powerful are graph neural networks?
\newblock In \emph{7th International Conference on Learning Representations,
  {ICLR} 2019, New Orleans, LA, USA, May 6-9, 2019}.

\bibitem[{Yang et~al.(2015)Yang, Yih, He, Gao, and
  Deng}]{DBLP:journals/corr/YangYHGD14a}
Bishan Yang, Wen{-}tau Yih, Xiaodong He, Jianfeng Gao, and Li~Deng. 2015.
\newblock Embedding entities and relations for learning and inference in
  knowledge bases.
\newblock In \emph{3rd International Conference on Learning Representations,
  {ICLR} 2015}.

\bibitem[{Zhang et~al.(2018)Zhang, Li, Mei, and Mao}]{DBLP:conf/www/ZhangLMM18}
Richong Zhang, Junpeng Li, Jiajie Mei, and Yongyi Mao. 2018.
\newblock Scalable instance reconstruction in knowledge bases via relatedness
  affiliated embedding.
\newblock In \emph{The World Wide Web Conference, {WWW} 2018}, pages
  1185--1194.

\bibitem[{Zhang et~al.(2020)Zhang, Zou, and Ma}]{Zhang2020Hyper-SAGNN}
Ruochi Zhang, Yuesong Zou, and Jian Ma. 2020.
\newblock Hyper-sagnn: a self-attention based graph neural network for
  hypergraphs.
\newblock In \emph{International Conference on Learning Representations}.

\end{thebibliography}

\clearpage
\appendix

\section{Further details on WD50K}
\label{app:dataset}
In contrast with Freebase which is no longer supported nor updated, we choose Wikidata as the source KG for our dataset since it has an active community and has seen contributions from various companies that merge their knowledge with it. Additionally, many new NLP tasks~\citep{Xiong2020Pretrained, DBLP:journals/corr/abs-1908-07690, DBLP:journals/corr/abs-1907-09361}, as well as datasets~\citep{DBLP:journals/corr/abs-1911-06136, DBLP:conf/emnlp/MesquitaCSMB19, dubey2019lc}, are using Wikidata as a reference KG.

\begin{figure}[!h]
    \centering
    \includegraphics[width=\linewidth]{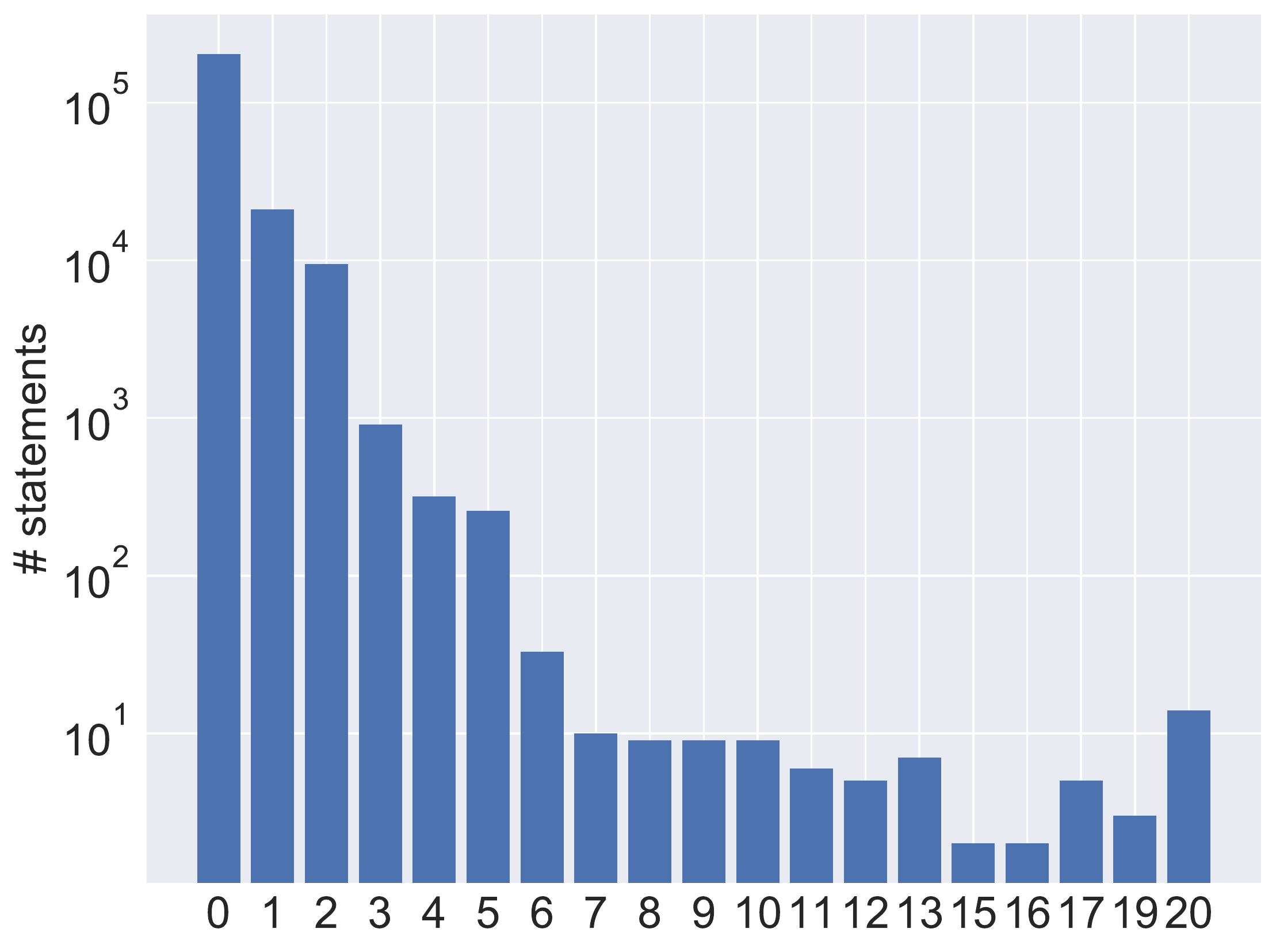}
    \caption{Number of qualifiers per statement}
    \label{fig:WD50K_quals_distr}
\end{figure}

\begin{figure}[!h]
    \centering
    \includegraphics[width=\linewidth]{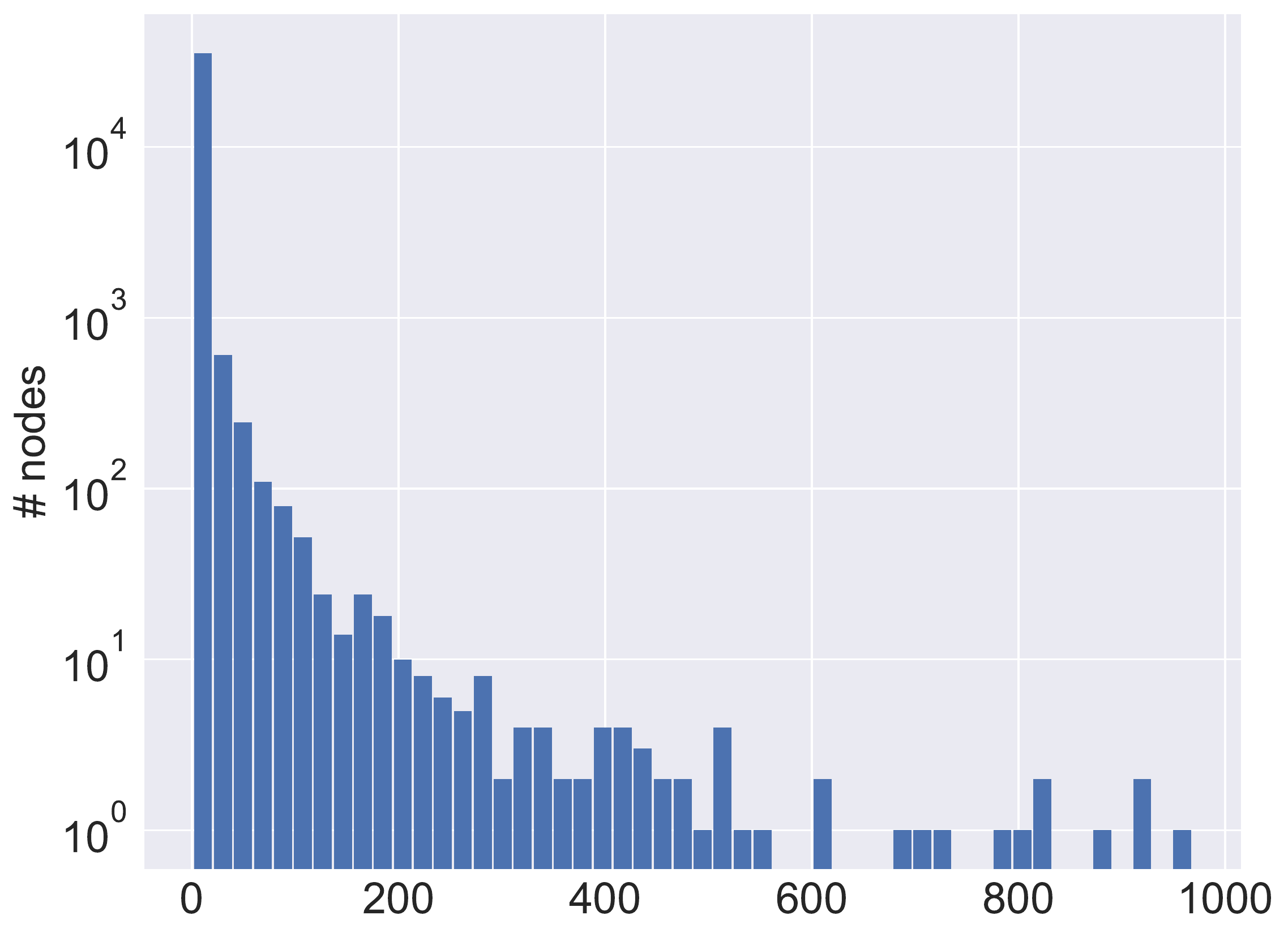}
    \caption{In-degree distribution}
    \label{fig:WD50K_indegree}
\end{figure}

The combined statistics of our dataset are presented in Table~\ref{tab:datasets}. 
WD50k consists of 47,156 entities, and 532 relations, amongst which 5,460 entities and 45 relations are found only within qualifier \textit{($q_p$, $q_e$)} pairs.
Fig.~\ref{fig:WD50K_quals_distr} illustrates how qualifiers are distributed among statements, i.e., 236,393 statements (99.9\%) contain up to five qualifiers whereas remaining 114 statements in a long tail contain up to 20 qualifiers.
Fig.~\ref{fig:WD50K_indegree} illustrates the in-degree distribution (with 50 bins, values higher than 1000 are omitted) of the WD50K graph structure where most of the nodes have in-degrees up to 200.

Recall that we augmented our dataset to reduce test set leakage by removing all instances from the train, and validation sets whose main triple $(\textit{s,\ p,\ o})$ can be found in the test instances (Sec.~\ref{sec:WD50K}). 
Another form of test leakage, as discovered in~\cite{toutanova2015observed}, may still persist in our dataset. 
To estimate this,  we count the instances in the test set whose main triple's  "direct" inverse $(\textit{o,\ p,\ s})$, or "semantic" inverse (based on the relation P1696 in Wikidata, i.e., inverse of) is present in the train set. This amounts to less than 4\% (1.6k out of 46k) instances in the test set.

\section{Sparse Representation}
\label{app:sparse}

\begin{figure}[h!]
    \centering
    \includegraphics[width=\linewidth]{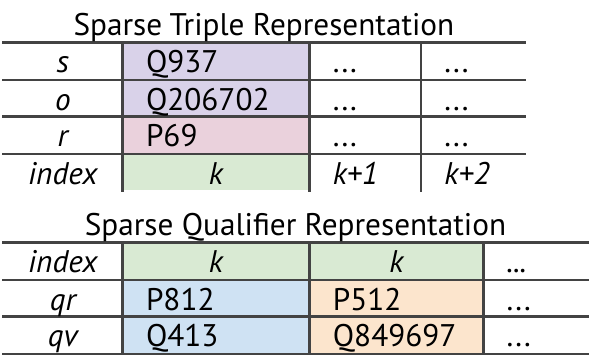}
    \caption{Sparse representation for hyper-relational facts. Each fact has a unique integer index $k$ which is shared between two COO matrices, i.e., the first one is for main triples, the second one is for qualifiers. Qualifiers that belong to the same fact share the index $k$.}
    \label{fig:sparse}
\end{figure}


Storing full adjacency matrices of large KGs is impractical due to $O(|\mathcal{V}|^2)$ memory consumption.
GNNs encourage using sparse matrix representations and adopting sparse matrices is shown~\citep{Cohen2020Scalable} to be scalable to graphs with millions of edges.
As illustrated in \autoref{fig:sparse}, we employ two sparse COO matrices to model hyper-relational KGs. 
The first COO matrix is of a standard format with rows containing indices of subjects, objects, and relations associated with the \emph{main triple} of a hyper-relational fact.  

In addition, we store index $k$ that uniquely identifies each fact.
The second COO matrix contains rows of qualifier relations $qr$ and entities $qe$ that are connected to their main triple (and the overall hyper-relational fact) through the index $k$, i.e., 
if a fact has several qualifiers those columns corresponding to the qualifiers of the fact will share the same index
$k$. 
The overall memory consumption is therefore $O(|\mathcal{E}|+|\mathcal{Q}|)$ and scales linearly to the total number of qualifiers $|\mathcal{Q}|$. 
Given that most open-domain KGs rarely qualify each fact, e.g., as of August 2019, out of 734M Wikidata statements approximately 128M (17.4\%) have at least one qualifier, this sparse qualifier representation  saves limited GPU memory.


\section{Hyperparameters}
\label{app:hyperparams}

We tuned the model (\textsc{StarE} encoder with Transformer decoder) on the validation set using the hyperparameters reported in Table~\ref{tab:hyperparams_search}.
Implementations of mult, ccorr, and rotate functions in $\phi_q$ and $\phi_r$ correspond to DistMult~\citep{DBLP:journals/corr/YangYHGD14a}, circular correlation~\citep{DBLP:conf/aaai/NickelRP16}, and RotatE~\citep{DBLP:conf/iclr/SunDNT19}, respectively.

\begin{table}[!h]
\centering
\caption{This table reports the major hyperparameters of our approach, and their corresponding bounds. Note that "Trf" refers to Transformers. Selected values are in \textbf{bold}.}
\label{tab:hyperparams_search}
\begin{tabular}{@{}lc@{}}
\toprule
Parameter & Value \\ \midrule
\textsc{StarE} layers & \{1, \textbf{2}\} \\
Embedding dim & \{100, \textbf{200}\} \\
Batch size & \{\textbf{128}, 256, 512\}  \\
Learning rate & \{\textbf{0.0001}, 0.0005, 0.001\} \\
$\phi_q$ & mult, ccorr, \textbf{rotate} \\
$\phi_r$ & mult, ccorr, \textbf{rotate} \\
$\gamma$ & \textbf{weighted sum} concat, mul \\
Weighted sum $\alpha$ & $[0.0, 1.0]$ step $0.1$ \\
Quals aggregation & \textbf{sum}, mean \\  
Trf layers & \{1, \textbf{2}\} \\
Trf hidden dim & \{256, \textbf{512}, 768\} \\
Trf heads & \{2, \textbf{4}\} \\
StarE dropout & \{0.1, 0.2, \textbf{0.3}\}\\
Trf dropout & \{\textbf{0.1}, 0.2, 0.3\}\\
Label smoothing & \{0.0, \textbf{0.1}\} \\
\bottomrule
\end{tabular}
\end{table}

The selected hyperparameters include two \textsc{StarE} layers, embedding dimension of 200, batch size of 128, Adam optimizer with 0.0001 learning rate and 0.1 label smoothing. $\phi_r$ and $\phi_q$ are rotate functions, $\gamma(\cdot)$ is a weighted sum function with $\alpha$ of 0.8, qualifiers are aggregated using a simple summation, and 0.3 dropout rate. 
We use 2-layer Transformer block with the hidden dimension of 512, and 4 attention heads with 0.1 dropout rate as our decoder.
For WD50K and JF17K datasets we set the maximum length of a hyper-relational fact to 15 (i.e., a statement can contain at most 6 qualifier pairs), and 7 for WikiPeople.

\textbf{Infrastructure and Parameters.}
We train all models on one Tesla V100 GPU. 
Due to a large number of parameters, owing to large trainable embedding matrices, it is advisable to a GPU with at least 12GB of VRAM.
Running \textsc{StarE} (H) + Transformer (H) models with the selected hyperparams on WD50K requires approximately 2 days to train and has 10.8M parameters\footnote{According to a built-in PyTorch counter.}; on JF17k the model has 7.1M parameters and takes about 10 hours to train; on WikiPeople the model has 8.2M parameters which we run for 500 epochs and takes about 4 days. 

StarE (H) + Transformer (H) models on reduced datasets: the model corresponding to WD50K (33) has 9M parameters and takes 20 hours to train while WD50K model has 6.8M parameters and takes about 9 hours to train. In case of WD50K (100), the model has 5M parameters and takes 5 hours to train.


\section{Decoders}
\label{app:decoders}

\begin{table*}[t]
\centering
\caption{Effect of different decoders on the link prediction task over WD50K, and its variations.}
\label{tab:decoders}
\adjustbox{max width=\linewidth}{
\begin{tabular}{@{}lcccccccccccc@{}}
\toprule
Dataset $\rightarrow$ & \multicolumn{3}{c}{\textbf{WD50K}} & \multicolumn{3}{c}{\textbf{WD50K (33)}} & \multicolumn{3}{c}{\textbf{WD50K (66)}} & \multicolumn{3}{c}{\textbf{WD50K (100)}} \\ \cmidrule(l){2-4} \cmidrule(l){5-7} \cmidrule(l){8-10} \cmidrule(l){11-13}
Method $\downarrow$ & MRR & H@1 & \multicolumn{1}{c}{H@10} & MRR & H@1 & \multicolumn{1}{c}{H@10} & MRR & H@1 & \multicolumn{1}{c}{H@10} & MRR & H@1 & H@10 \\ \midrule
\textsc{StarE} + Trf & \textbf{0.349} & \textbf{0.271} &\textbf{0.496} & \textbf{0.331} & \textbf{0.268} & 0.451 & \textbf{0.481} & \textbf{0.420} & 0.594 & \textbf{0.654} & \textbf{0.588} & \textbf{0.777} \\
\textsc{StarE} + ConvE & 0.341 & 0.260 & \textbf{0.496} & 0.323 & 0.254 & \textbf{0.456} & 0.460 & 0.392 & 0.590 & 0.627 & 0.550 & 0.772 \\
\textsc{StarE} + ConvKB & 0.323 & 0.241 & 0.479 & 0.316 & 0.247 & 0.448 & 0.448 & 0.377 & 0.584 & 0.621 & 0.544 & 0.763 \\
\textsc{StarE} + MskTrf & 0.341 & 0.262 & 0.489 & 0.324 & 0.260 & 0.446 & 0.479 & 0.417 & \textbf{0.595} & 0.649 & 0.579 & 0.774 \\
\bottomrule
\end{tabular}}
\end{table*}

\begin{figure*}[t]
    \centering
    \includegraphics[width=\linewidth]{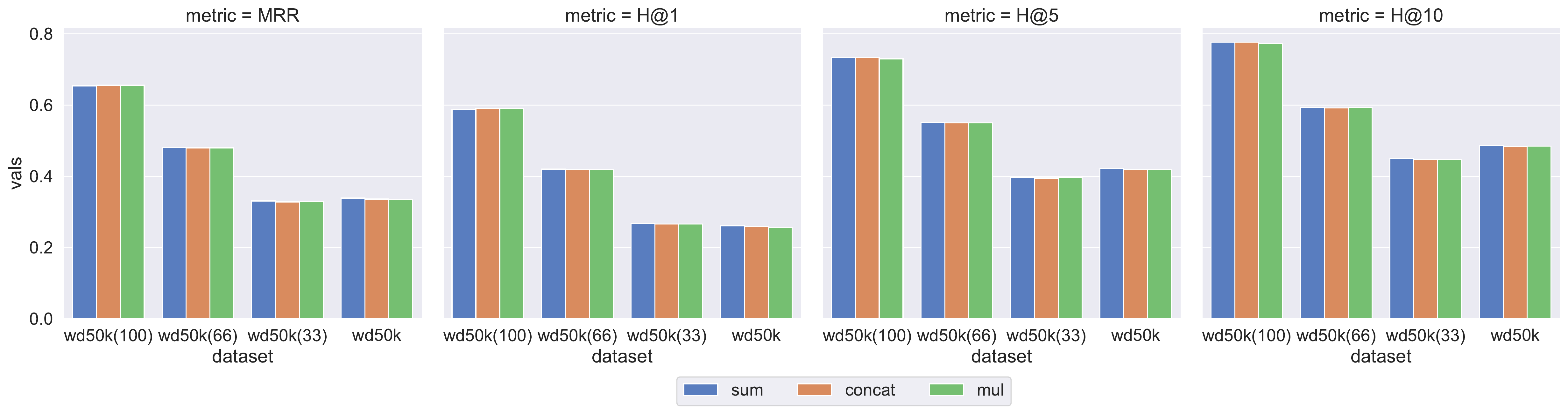}
    \caption{Gamma experiment.}
    \label{fig:WD50K_gamma1}
\end{figure*}

As an additional experiment, we pair \textsc{StarE} with different decoders and evaluate them over
WD50K datasets. 
Along with the main reported model denoted as \textbf{StarE + Trf}, we implemented two CNN-based decoders and another Transformer-based decoder. 
All models are trained with the same encoder hyperparameters as chosen in the main reported model.

\textbf{StarE + ConvE} relies on the ConvE~\citep{DBLP:conf/aaai/DettmersMS018}-like decoder but expanded for statements with qualifiers. 
Given a query \textit{(s, r, \{$(qr_{i}$,  $qv_{i})$, ... \})}, we stack entities and relations embeddings row-wise and reshape the tensor into an \emph{image} of size $H \times W $. 
For instance, for a statement with 6 qualifier pairs, i.e., query length of 14, and an embedding size of 200, we obtain \emph{images} of size $40\times70$. 
We then apply a 2D convolutional layer with a $7 \times 7$ kernel for each image, apply ReLU, flatten the resulting tensor, and pass it through a fully-connected layer. 
We used 200 filters and the learning rate was set to 0.001. 

\textbf{StarE + ConvKB} is based on the ConvKB~\citep{DBLP:conf/naacl/NguyenNNP18}-like decoder adjusted for statements with qualifiers.
Given a query \textit{(s, r, \{$(qr_{i}$,  $qv_{i})$, ... \})}, we stack entities and relations embeddings row-wise and apply a 2D convolutional layer with a $L_Q \times 7$ kernel, e.g., for queries of length 14 the kernel size is $14 \times 7$. 
We then apply ReLU, flatten the resulting tensor, and pass it through a fully-connected layer.
We used 200 filtersand the learning rate was set to 0.001.


\textbf{StarE + MskTrf} denotes a Transformer decoder with an explicit \texttt{[MASK]} token at the object position of each query.
Given a query \textit{(s, r, \{$(qr_{i}$,  $qv_{i})$, ... \})}, we extract relevant entities and relation embeddings and insert the \texttt{[MASK]} token, transforming it into \textit{(s, r, \texttt{[MASK]}, \{$(qr_{i}$,  $qv_{i})$, ... \})}.
We then pass it through the Transformer layers and retrieve the  representation of the \texttt{[MASK]} token. 
Finally, the token representation is passed through a fully-connected layer.
We trained the model with 0.0001 as the learning rate.

Table~\ref{tab:decoders} reports link prediction results on a variety of WD50K datasets with with different decoders. 
The default \textbf{StarE + Trf} decoder generally attains superior results with biggest gains along H@1 metric.

\section{Relation-Qualifiers Aggregation}
In this experiment, we measure the impact of the choice of $\gamma(\cdot)$ function which is used for aggregating representations of a relation and its qualifiers (see Eq.~\ref{eq:stare}). To evaluate its impact we use \textsc{StarE} (H) + Transformer (H) models, on four WD50K datasets using three functions, i.e., concatenation $[\mathbf{h}_r,\mathbf{h}_{q}]$, element-wise multiplication $\mathbf{h}_r \odot \mathbf{h}_{q}$, and weighted sum $\alpha \odot \mathbf{h}_r + (1-\alpha) \odot \mathbf{h}_{q}$ where $\alpha$ is fixed to 0.8.


The results are presented in Fig.\ref{fig:WD50K_gamma1}.
We find that all the three settings have similar performance indicating model's stability with respect to the choice of $\gamma(\cdot)$ function.

\end{document}